\documentclass{article}
\usepackage[preprint]{neurips_2026}
\workshoptitle{Sim2Science}

\usepackage[utf8]{inputenc}
\usepackage[T1]{fontenc}
\usepackage{hyperref}
\usepackage{url}
\usepackage{booktabs}
\usepackage{graphicx}
\usepackage{float}
\usepackage{amsfonts}
\usepackage{amsmath}
\usepackage{nicefrac}
\usepackage{microtype}
\usepackage{xcolor}

\usepackage{svg}

\title{Component-Level Evaluation of Adaptive PINN Training for CFD-Oriented Crystal Growth Simulation}

\author{
Niruta Chapagain\\
University of Europe for Applied Sciences\\
Potsdam, Germany\\
\texttt{nirutachapagain825@gmail.com}
\And
Rohit Raj\\
Vienna, Austria\\
\texttt{rajrohit403@gmail.com}
\And
Bertwin Kurisinkal Shine\\
Technische Universität Chemnitz\\
Chemnitz, Germany\\
\texttt{bertwinkurisinkalshine@gmail.com}
\And
Aditya A S\\
quasi digital\\
Berlin, Germany\\
\texttt{aditya@quasi.digital}
}

\begin{document}
\raggedbottom

\maketitle
\begin{abstract}
    Physics-informed neural network (PINN) training minimizes a weighted combination of partial differential equation (PDE), boundary-condition, and initial-condition losses. Because adaptive methods modify these weights during training, their weighted total losses are not always directly comparable. We compare fixed-weight PINN, gradient-normalized PINN, and a rule-based adaptive controller under matched settings on a heat-equation benchmark and a simplified Czochralski-oriented thermal-fluid problem. In the crystal-growth MLP experiment, adaptive control reduced the PDE residual from the order of $10^{-5}$ to $10^{-6}$, while the boundary-condition loss increased from the order of $10^{-5}$ to $10^{-2}$. On the heat-equation benchmark, GNPINN achieved the lowest relative $L_2$ field error ($0.054$), whereas AgenticPINN obtained the smallest PDE residual but a relative $L_2$ error of $1.368$. Gaussian-process surrogates were additionally evaluated using case-wise holdout tests on corrected Czochralski CFD parameter sweeps. The temperature-field error for the temperature sweep was approximately $6\%$, whereas the axial-velocity error for the crystal-rotation sweep was approximately $42\%$. These findings show that adaptive control can improve equation satisfaction while weakening other physical constraints. PINN training should therefore be evaluated using separate PDE, boundary-condition, and solution-error metrics rather than weighted total loss alone.
\end{abstract}

%--------------------------------------------------------------%
\section{Introduction}

Czochralski silicon crystal growth %is a physically coupled process involving heat transfer, melt flow, pressure variation, and %boundary conditions.
couples heat transfer, melt convection, pressure variations and rotating boundaries, and resulting thermal and flow fields largely determine crystal quality. %Classical 
Computational fluid dynamics (CFD) can represent such behaviour effectively, but repeated high-fidelity simulations are expensive when many process parameters such as sweeping over pull rate, crystal rotations, thermal boundaries conditions etc., are needed to be studied \citep{jin2013czfluidflow,enders2022czthermal}. The cost is therefore not one simulation but the sweep, which is what motivates cheaper learned emulators.

PINNs are attractive for this type of problem because they include governing equations directly in the training objective instead of relying only on labelled data \citep{raissi2019physics,karniadakis2021physics}. However, PINN training is not automatically stable. Prior studies show that optimization stiffness, gradient imbalance, and competing loss terms can lead to failure even when the PDE is known \citep{wang2021gradientpathologies,krishnapriyan2021failure}. %This is especially important for CFD-oriented learning because a low total loss may hide poor boundary satisfaction or weak physical consistency.
These failures are usually diagnosed after the fact, by inspecting which part of the physics the trained model violates, which raises the question of whether standard reporting practice makes that inspection possible at all.

A central difficulty in evaluating PINNs is that optimization success does not necessarily imply physical solution accuracy. PINN objectives combine PDE, boundary-condition, and initial-condition losses, and adaptive strategies modify their relative influence during training. Consequently, two methods may achieve similar or even lower weighted objectives while satisfying different physical constraints. The weighted training objective should therefore not automatically be interpreted as a common cross-method accuracy metric.

For a training strategy $m$, let
\begin{equation}
    \mathcal{L}^{(m)} = \sum_k \lambda_k^{(m)}\mathcal{L}_k^{(m)},
\end{equation}
where $k$ indexes physical loss components. If two strategies use different weights, $\lambda_k^{(a)}\neq\lambda_k^{(b)}$, then $\mathcal{L}^{(a)}<\mathcal{L}^{(b)}$ does not imply the same ordering of individual constraint violations or external solution error. This motivates evaluating adaptive PINNs using unweighted component losses and reference solution errors where available.

We study this distinction using a fixed-weight PINN, gradient-normalized training, and a rule-based adaptive controller. A one-dimensional heat equation provides analytical ground truth, while a simplified Czochralski-inspired thermal-fluid problem tests whether the same residual--constraint trade-off appears in a multi-physics setting. Case-wise CFD surrogate experiments further examine whether aggregate prediction quality can conceal field-specific errors.

Our contributions are threefold: (i) We clarify why adaptively weighted PINN objectives should be treated as training quantities rather than directly comparable cross-method accuracy measures, motivating component-resolved evaluation; (ii) in matched experiments, we show that substantially lower PDE residual can coexist with worse physical-constraint satisfaction and, on an analytical heat-equation benchmark, substantially larger solution error; and (iii) we extend the component-resolved evaluation perspective to Czochralski CFD surrogates, showing that predictive error can vary substantially across physical fields and operating-parameter sweeps.

%-----------------------------------------------------------------%
%\section{Method}
\section{Problem setting} \label{sec:problem_setting}
%%---------------------------%%

\subsection{Czochralski growth and CFD reference data}
\label{sec:cfd_reference}

%Czochralski silicon crystal growth involves coupled heat transfer, melt convection, pressure variation, and rotating boundaries, making repeated CFD simulations costly when process parameters are swept. We therefore use CFD solutions as reference data for surrogate modelling.

The reference cases were generated with ANSYS Fluent 2026 R1 as steady, laminar, two-dimensional axisymmetric simulations of the silicon melt. The model solves the incompressible Navier--Stokes and energy equations with Boussinesq buoyancy and azimuthal velocity induced by crystal and crucible rotation. Rotating walls satisfy $u_\theta=\omega r$; the free surface is non-penetrating and slip; and prescribed thermal boundary conditions are applied at the free surface and crucible bottom. A mesh-independence study established a common grid of 8181 nodes. Because the rotational Reynolds numbers are large, these simulations are treated as controlled reference cases rather than fully resolved industrial melt-flow predictions.

We consider three one-parameter sweeps: hot-wall temperature $T_{\mathrm{hot}}\in[1730,1785]\,\mathrm{K}$, crucible angular velocity $\omega_c\in[-10,-1]\,\mathrm{rpm}$, and crystal angular velocity $\omega_s\in[4,20]\,\mathrm{rpm}$ \citep{shine2026czsiliconmelt}. Each case provides axisymmetric fields over $(r,z)$, and the surrogate learns $(r,z,s) \mapsto (u_r,u_z,u_\theta,p,T)$, where $s$ is the active process parameter.

One complete operating condition is withheld from each sweep for testing: $1780\,\mathrm{K}$, $-4\,\mathrm{rpm}$, and $7\,\mathrm{rpm}$ for the temperature, crucible-rotation, and crystal-rotation sweeps, respectively. This case-wise holdout evaluates generalization to an unseen operating condition rather than interpolation among randomly split spatial points ~\citep{shine2026czsiliconmelt}.

For the surrogate experiments, Gaussian-process regression (GPR) models are
trained on the available operating conditions and evaluated on the withheld
case. Performance is reported separately for each output field using
normalized RMSE (NRMSE), defined as RMSE divided by the standard deviation
of the corresponding target field. Implementation and kernel details are
provided in Appendix~\ref{app:gpr_details}.

%We use CFD sweeps over \todo{BE todo: do mention parameter in full first before mention its short script}[PARAMETERS] as supervised reference data \citep{bertwinCZTempChange,bertwinCZCrucibleSweep,bertwinCZCrystalSweep}. Each case is an axisymmetric steady-state solution on \todo{BE todo}[DOMAIN], and the sweeps vary \todo{BE todo}[PARAM RANGES]. The surrogate maps spatial coordinates and a process
%parameter to the field variables,
%\begin{equation} \label{eq:surrogate}
   % (r,z,s) \mapsto (u_r,u_z,u_{\theta},p,T),
%\end{equation}

%and is trained under a case-wise holdout in which one full operating condition is unseen. This tests generalization across process settings rather than interpolation within a randomly split point cloud, and it lets surrogate error be reported per field rather than pooled.
%%-----------------------------------------%%

\subsection{Governing equations and training objective}
For a neural approximation $\hat{u}_{\theta}$, the physics-informed objective is
\begin{equation} \label{eq:objective}
    \mathcal{L}(\theta)
    = \lambda_{\mathrm{PDE}}\mathcal{L}_{\mathrm{PDE}}
    + \lambda_{\mathrm{BC/IC}}\mathcal{L}_{\mathrm{BC/IC}}
\end{equation}

where $\mathcal{L}_{\mathrm{PDE}}$ is the equation residual and $\mathcal{L}_{\mathrm{BC/IC}}$ measure boundary and/or initial-condition error. The strategies of Section~\ref{sec:training_strategies}  differ in how the PDE and BC/IC contributions are balanced during optimization. We therefore record the physical loss components separately throughout training. Reporting $\mathcal{L}(\theta)$ alone is what this paper examines: the physical loss components are recorded separately throughout.

% We study two physics settings. The 1D heat equation,
% \begin{equation} \label{eq:heat}
%     \frac{\partial T}{\partial t} = \alpha \frac{\partial^2 T}{\partial x^2}, \
% \end{equation}

% on $x\in[0,1]$ and $t\in[0,1]$. Homogeneous Dirichlet boundary conditions are imposed at both ends of the spatial domain:

% \begin{equation}
%     T(0,t)=0, \qquad T(1,t)=0.
% \end{equation}

% The initial temperature distribution is

% \begin{equation}    
%     T(x,0)=\sin(\pi x).
% \end{equation}

% For these conditions, the analytical solution is

% \begin{equation}
%     T_{\mathrm{exact}}(x,t) = \sin(\pi x)\exp(-\alpha\pi^2t).
% \end{equation}
We consider the 1D heat equation on $(x,t)\in[0,1]^2$,
\begin{equation}\label{eq:heat}
    \frac{\partial T}{\partial t}
    = \alpha\frac{\partial^2T}{\partial x^2},
    \quad
    T(0,t)=T(1,t)=0,
    \quad
    T(x,0)=\sin(\pi x),
\end{equation}
whose analytical solution is $T_{\mathrm{exact}}(x,t)=\sin(\pi x)\exp(-\alpha\pi^2t)$.

The reported heat-equation experiments use $\alpha=1$. This is the only physics setting in the study for which an analytical reference solution is available. It therefore allows the trained field to be evaluated directly, in addition to comparing the individual loss components.

The simplified crystal-growth PINN benchmark and the CFD surrogate dataset serve distinct roles: the former evaluates physics-informed optimization behavior, whereas the latter evaluates case-wise surrogate generalization over CFD-generated parameter sweeps.

%-----------------------------------------------%
\section{Training strategies}
\label{sec:strategies}
\label{sec:training_strategies}

\paragraph{Baseline PINN.}
The baseline minimizes $\mathcal{L}_{\mathrm{PDE}}+\mathcal{L}_{\mathrm{BC/IC}}$ with fixed unit weights. New collocation points are sampled at every epoch, and Adam updates the network from the gradient of the combined loss.

\paragraph{GNPINN.}
GNPINN computes the PDE and BC/IC gradients separately. For each parameter tensor, every gradient with norm greater than $10^{-8}$ is divided by its Euclidean norm. The two normalized gradients are then added and supplied to Adam. This limits domination by the loss component with the larger raw gradient magnitude.

\paragraph{AgenticPINN (adaptive controller).}
AgenticPINN is a project-specific, rule-based adaptive controller, rather than an LLM-based agent. It observes the global gradient norms of the PDE and BC/IC losses and adjusts their weights during training. Starting from $\lambda_{\mathrm{PDE}}=\lambda_{\mathrm{BC/IC}}=1$, the controller uses exponential moving averages with momentum $m=0.9$. After 30 warm-up epochs, weights are updated every five epochs using

% \begin{equation}
% \begin{aligned}
%     a&=\operatorname{clip}\left(
%     \sqrt{\frac{\bar g_{\mathrm{PDE}}+10^{-12}}
%     {\bar g_{\mathrm{BC/IC}}+10^{-12}}},\,0.8,\,1.25\right), \\
%     \lambda_{\mathrm{PDE}} &\leftarrow
%     \operatorname{clip}\left(\lambda_{\mathrm{PDE}}/a,10^{-2},10^2\right),\\
%     \lambda_{\mathrm{BC/IC}} &\leftarrow
%     \operatorname{clip}\left(\lambda_{\mathrm{BC/IC}}a,10^{-2},10^2\right).
% \end{aligned}
% \end{equation}
\begin{equation}
\begin{aligned}
a&=\operatorname{clip}\!\left(
\sqrt{\frac{\bar g_{\mathrm{PDE}}+10^{-12}}
{\bar g_{\mathrm{BC/IC}}+10^{-12}}},\,0.8,\,1.25\right),\\
(\lambda_{\mathrm{PDE}},\lambda_{\mathrm{BC/IC}})
&\leftarrow
\Bigl(
\operatorname{clip}(\lambda_{\mathrm{PDE}}/a,10^{-2},10^2),\,
\operatorname{clip}(\lambda_{\mathrm{BC/IC}}a,10^{-2},10^2)
\Bigr).
\end{aligned}
\end{equation}

After 200 epochs, the controller also compares the mean loss of two consecutive 100-epoch windows. If relative improvement is below $1\%$, the learning rate is multiplied by $0.7$, with a minimum of $10^{-6}$. Within each comparison, all three strategies use the same architecture, optimizer, sampling budget, initialization seed, and initial learning rate.

%------------------------------------%
\section{Results} 
\label{sec:results}

\subsection{Component-resolved PINN comparison}
Table~\ref{tab:pinn_quantitative} reports the final PDE and boundary and/or initial-condition losses separately. For the heat-equation benchmark, solution accuracy is additionally measured against the analytical solution using the relative $L_2$ error. %The experiments used matched model and sampling settings within each problem, so the component values reveal how each training strategy distributed its error. %The weighted total is included for completeness, but it should not be treated as a common accuracy scale when a controller changes the loss weights during training.

\begin{table}[h!]
\centering
\small
\caption{Final PINN losses under matched settings. PDE and BC/IC losses are reported separately because the adaptive strategy changes their relative weighting during training. Relative $L_2$ error is available only for the heat-equation benchmark.}
\label{tab:pinn_quantitative}
\resizebox{\textwidth}{!}{%
\begin{tabular}{llrrrr}
\toprule
Problem & Trainer & PDE loss & BC/IC loss & Weighted total & Relative $L_2$ error \\
\midrule
Heat-equation MLP & Baseline PINN & $3.96\times10^{-3}$ & $1.15\times10^{-3}$ & $5.11\times10^{-3}$ & $8.43\times10^{-2}$ \\
Heat-equation MLP & GNPINN & $9.74\times10^{-3}$ & $\mathbf{9.05\times10^{-4}}$ & $1.06\times10^{-2}$ & $\mathbf{5.41\times10^{-2}}$ \\
Heat-equation MLP & AgenticPINN & $\mathbf{8.22\times10^{-7}}$ & $2.00\times10^{-1}$ & $2.08\times10^{-3}$ & $1.368$ \\
\midrule
Crystal-growth MLP & Baseline PINN & $9.26\times10^{-6}$ & $1.63\times10^{-5}$ & $2.55\times10^{-5}$ & -- \\
Crystal-growth MLP & GNPINN & $1.61\times10^{-5}$ & $1.01\times10^{-4}$ & $1.17\times10^{-4}$ & -- \\
Crystal-growth MLP & AgenticPINN & $\mathbf{1.14\times10^{-6}}$ & $1.29\times10^{-2}$ & $2.43\times10^{-4}$ & -- \\
\bottomrule
\end{tabular}}
\end{table}

\paragraph{Heat-equation benchmark.}
The analytical benchmark exposes a clear disagreement between residual minimization and solution accuracy. AgenticPINN reduced the PDE residual by approximately $4.82\times10^{3}$ relative to the baseline, but its BC/IC loss increased by approximately $173$ times. Despite having the smallest PDE residual, it produced a relative $L_2$ error of $1.368$. GNPINN, in contrast, obtained the lowest relative $L_2$ error ($5.41\times10^{-2}$), even though its PDE residual was substantially larger. Thus, lower PDE residual alone did not imply a more accurate solution field.

\paragraph{Simplified crystal-growth.}
The same trade-off appears in the thermal-fluid PINN. AgenticPINN reduced the PDE residual by approximately $8.1$ times relative to the baseline, but its BC/IC loss increased by roughly $792$ times. The baseline therefore provided a more balanced satisfaction of the governing equations and constraints, while GNPINN did not improve either component in this experiment. Because no analytical solution is available for this benchmark, these results are interpreted as constraint-satisfaction diagnostics rather than direct measures of field accuracy.

\begin{figure}[t]
    \centering
    \includegraphics[width=0.9\textwidth]{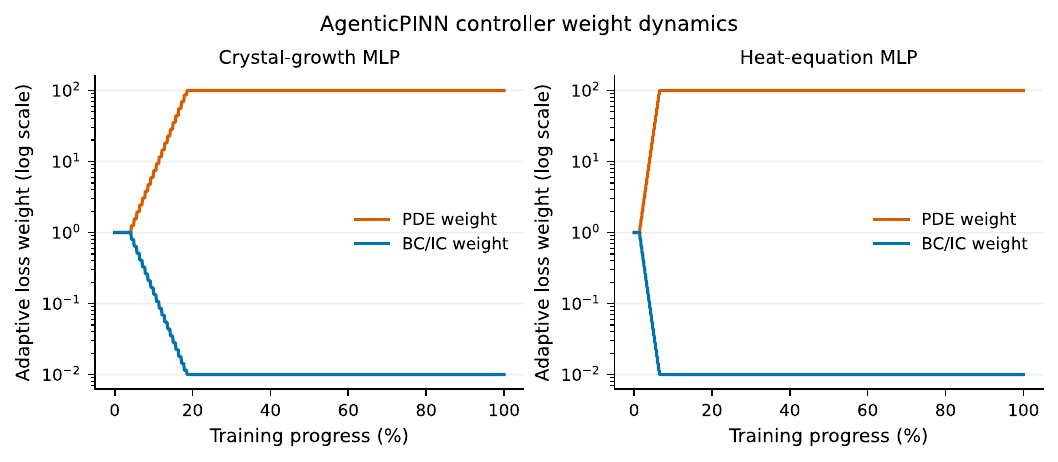}
    \caption{Adaptive loss-weight dynamics for the simplified crystal-growth (left) and heat-equation (right) PINNs. In both experiments, $\lambda_{\mathrm{PDE}}$ increases to its upper clipping bound while the constraint weight decreases to its lower bound, producing a $10^4$ disparity between the two loss weights.}
    \label{fig:controller_dynamics}
\end{figure}

\paragraph{Controller dynamics.}
Figure~\ref{fig:controller_dynamics} provides a mechanistic explanation for the observed trade-off. In both PINN experiments, the adaptive controller drives the PDE weight toward its upper bound ($\lambda_{\mathrm{PDE}}=10^2$) while the constraint weight approaches its lower bound ($\lambda_{\mathrm{C}}=10^{-2}$). The resulting $10^4$ weight ratio strongly prioritizes equation-residual reduction and suppresses the contribution of boundary and initial-condition errors to the training objective. This explains how the adaptive model can attain a small weighted objective and PDE residual while retaining a large constraint error and, on the heat benchmark, poor solution accuracy.

%The heat-equation MLP showed a more extreme version of the same pattern. AgenticPINN reduced the PDE residual by approximately $4.82\times10^{3}$ relative to the baseline, but its BC/IC loss increased by about $173$ times. GNPINN instead obtained the lowest BC/IC component, while retaining a larger PDE residual. These outcomes demonstrate why a single final loss cannot identify which physical constraint has been satisfied.

%Comparison with the analytical heat-equation solution confirms this distinction. On a fixed $201\times201$ evaluation grid, GNPINN achieved the lowest RMSE ($8.785\times10^{-3}$) and relative $L_2$ error ($5.41\times10^{-2}$), followed by the baseline PINN with RMSE $1.368\times10^{-2}$ and relative $L_2$ error $8.43\times10^{-2}$. AgenticPINN produced the smallest PDE residual but a substantially larger RMSE ($2.220\times10^{-1}$) and relative $L_2$ error ($1.368$). The analytical benchmark therefore shows directly that minimizing the PDE residual alone does not guarantee the most accurate solution field; boundary and initial-condition satisfaction remain essential.

\begin{figure}[t]
    \centering
    \includegraphics[width=0.9\textwidth]{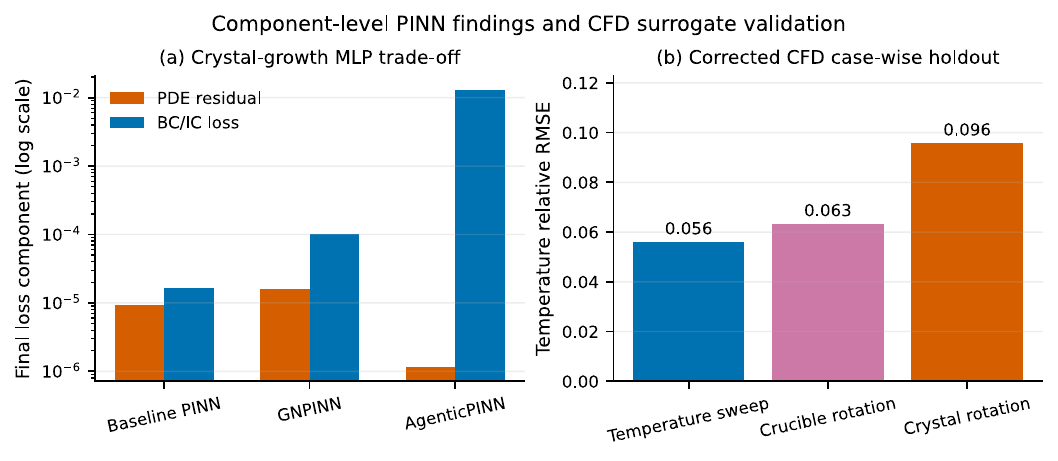}
    \caption{Component-level PINN and CFD-surrogate results. left: Final PDE and constraint losses for the simplified crystal-growth PINN, shown on a logarithmic scale. AgenticPINN obtains the smallest PDE residual but the largest constraint loss. right: Temperature-field normalized RMSE for case-wise holdouts from the three CFD parameter sweeps.}
    \label{fig:results_overview}
\end{figure}

%Figure~\ref{fig:results_overview}(a) makes the component trade-off visible on a logarithmic scale. AgenticPINN achieves the smallest crystal-growth PDE residual but the largest boundary error, whereas the baseline remains comparatively balanced. The result supports component-resolved reporting rather than a claim that one controller is universally superior.

\subsection{CFD surrogate evaluation} 
Gaussian-process surrogates were evaluated using one complete operating condition withheld from each CFD parameter sweep. Errors are reported field-wise rather than pooled across outputs. For temperature, normalized RMSE was $5.61\%$, $6.34\%$, and $9.58\%$ for the temperature, crucible-rotation, and crystal-rotation sweeps, respectively (Fig.~\ref{fig:results_overview} right). The crystal-rotation sweep was substantially harder for the meridional velocity components, with normalized RMSE of $22.95\%$ for $u_r$ and $41.46\%$ for $u_z$, while the corresponding swirl-velocity error remained only $2.93\%$. Thus, surrogate accuracy depends strongly on both the operating-parameter sweep and the physical field being predicted.

% \begin{figure}[t]
%     \centering
%     \includegraphics[width=\textwidth]{figures/gpr_uncertainty_error_correlation.svg}
%     \caption{Correlation between absolute holdout error and Gaussian-process predictive uncertainty. Positive correlation indicates that larger uncertainty tends to accompany larger error; values near zero or below zero indicate weak calibration for that target and sweep.}
%     \label{fig:gpr_uncertainty}
% \end{figure}

%Predictive uncertainty was informative for some targets but not uniformly so (Fig.~\ref{fig:gpr_uncertainty}). For the temperature sweep, the strongest error--uncertainty correlations occurred for $u_z$ ($0.48$) and $u_r$ ($0.42$). In contrast, the crystal-rotation sweep showed weak correlations for several outputs and slightly negative correlations for pressure and temperature. Predictive uncertainty should therefore be interpreted as a diagnostic signal whose usefulness depends on the field and operating regime, rather than as an automatically reliable indicator of prediction error.

%\subsection{Main result}

%Across the PINN and CFD-surrogate experiments, the recurring result is that aggregate performance can hide output-specific failure. Adaptive weighting can strongly improve an equation residual while degrading a boundary constraint, and a surrogate can model temperature accurately while missing a change in a velocity component. The central reporting recommendation is therefore to retain component-level losses, field-wise errors, and uncertainty diagnostics alongside any aggregate score.

\section{Discussion and Conclusion}

%The main contribution of this work is a focused comparison of fixed, gradient-normalized, and controller-guided PINN training for a CFD-oriented crystal-growth problem. AgenticPINN is not claimed as a universal algorithm. It is a project-specific adaptive controller used to study how feedback-based loss adjustment changes training behaviour.

%The results show that adaptive control is useful but conditional. It can improve PDE residuals in selected cases, yet it may also move error toward boundary or initial-condition terms. For scientific machine learning, this means that total loss alone is not enough. Separate reporting of PDE, boundary, initial-condition, and surrogate errors gives a more honest view of model behaviour.

%The study is limited to simplified PDE settings and local-compute experiments. Future work should connect adaptive PINN training more directly with larger CFD datasets, uncertainty-aware surrogate models, and validation against high-fidelity solvers.
The experiments distinguish three quantities that need not agree in PINN evaluation: the weighted training objective, satisfaction of individual physical constraints, and solution accuracy. On the heat equation, AgenticPINN reduced the PDE residual by several orders of magnitude but produced a substantially larger relative $L_2$ error, coinciding with a large increase in constraint loss. The simplified crystal-growth PINN showed the same residual--constraint trade-off, although no analytical solution is available for direct field-error evaluation. 

The CFD surrogate results reveal a similar need for component-level evaluation: temperature was predicted comparatively accurately, whereas some velocity components, particularly under crystal rotation, showed much larger errors. %Predictive uncertainty was also informative for some field--sweep combinations but weak for others. 

Overall, adaptively weighted objectives should be treated as optimization quantities rather than common cross-method accuracy metrics. We therefore recommend reporting unweighted physical-loss components, external solution error where available, and field-wise surrogate errors. Future work should test these findings across additional PDEs, random initializations, and higher-fidelity crystal-growth models.

\paragraph{Limitations.} The PINN comparisons use a single matched initialization and two physics settings, so the observed optimization behavior should not be interpreted as universal across adaptive weighting methods or PDEs. The simplified crystal-growth PINN has no analytical reference solution, and its component losses therefore diagnose constraint satisfaction rather than field accuracy. In addition, the CFD cases are steady, laminar, and axisymmetric reference simulations rather than fully resolved industrial Czochralski flow. Future work should test the findings across additional random initializations, PDEs, adaptive weighting strategies, and higher-fidelity crystal-growth models.

\bibliographystyle{plainnat}
\bibliography{sim2science_references}

\clearpage

\clearpage

\appendix
\section{Post-processed field comparisons}
\label{app:field_comparisons}

This appendix compares the analytical heat-equation solution with the fields learned by the three PINN training strategies under the final matched experiment settings. It also presents qualitative predictions from the simplified crystal-growth problem; these are not direct comparisons against high-fidelity CFD ground truth. A common color scale is used across the methods for each physical variable, allowing the field structures to be compared directly. The predictions should nevertheless be interpreted together with the component-level losses reported in the main paper.

\begin{figure}[h!]
    \centering
    \includegraphics[width=\textwidth]{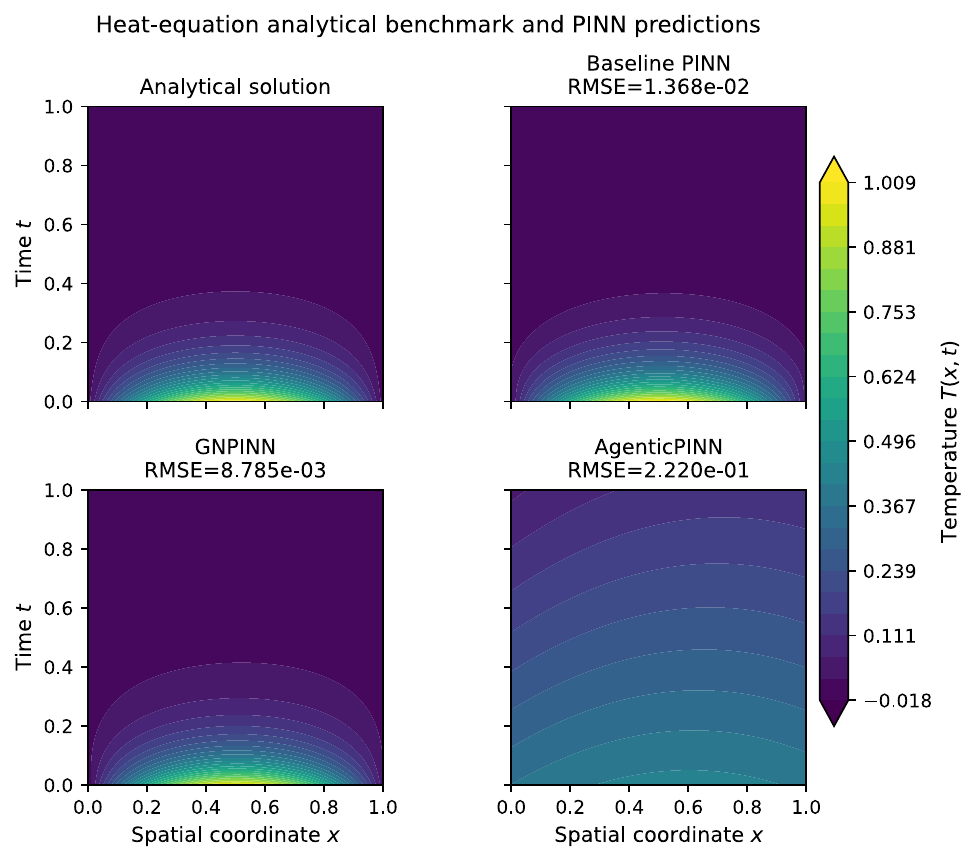}
    \caption{Analytical heat-equation solution and temperature predictions from the Baseline PINN, GNPINN, and AgenticPINN MLP runs. On the common evaluation grid, the corresponding relative $L_2$ errors are $8.43\times10^{-2}$, $5.41\times10^{-2}$, and $1.368$. All four panels use the same axes and color scale. Although AgenticPINN has the smallest PDE residual, its larger field error is consistent with its substantially larger boundary and initial-condition loss.}
    \label{fig:appendix_heat_fields}
\end{figure}

\begin{figure}[h!]
    \centering
    \includegraphics[width=\textwidth]{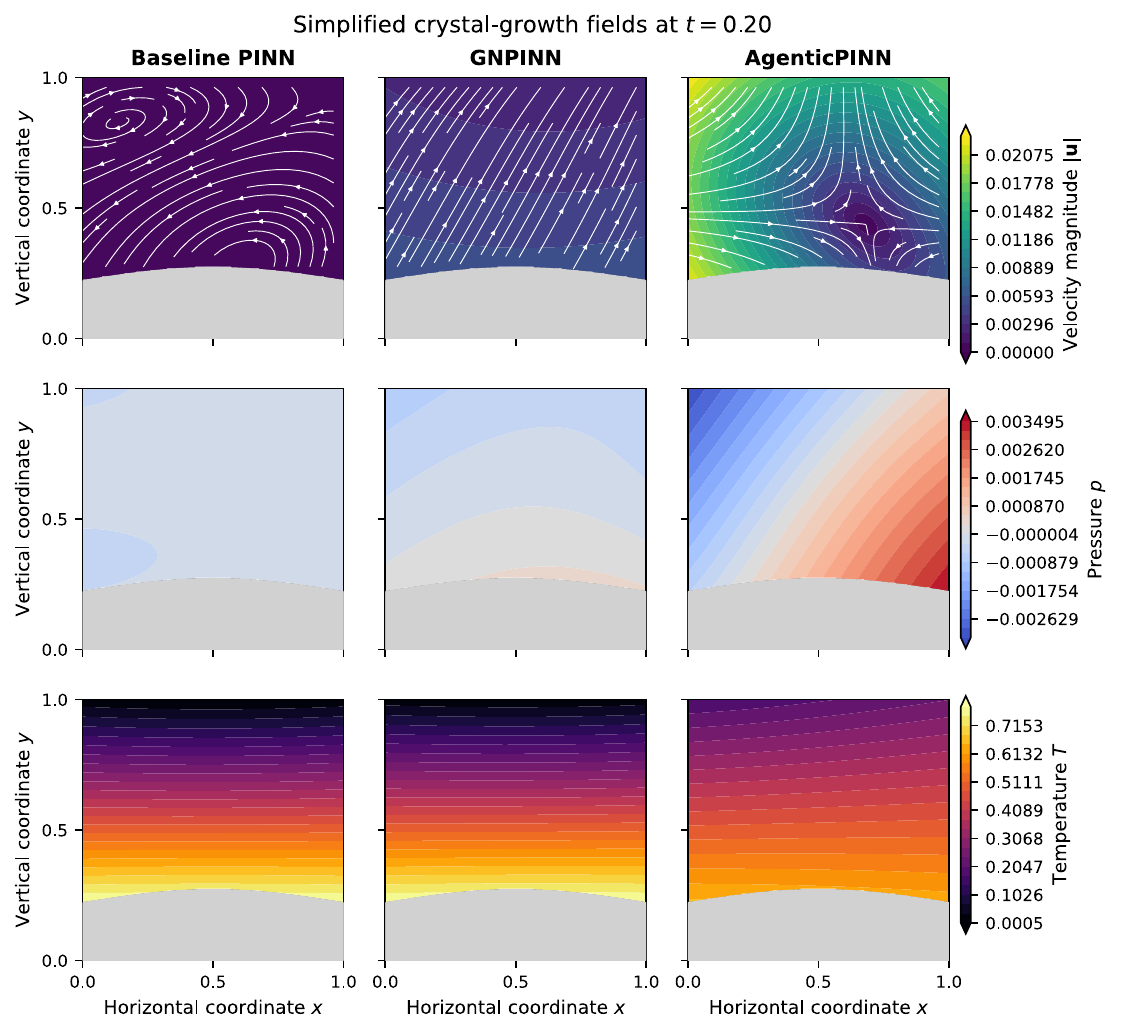}
    \caption{Post-processed velocity, pressure, and temperature predictions at $t=0.20$ for the simplified crystal-growth MLP experiment. Rows correspond to Baseline PINN, GNPINN, and AgenticPINN. The methods produce noticeably different velocity and pressure structures, while their temperature fields are more similar. These panels support the component-level interpretation of the training results but do not constitute validation against an industrial CFD solution.}
    \label{fig:appendix_crystal_fields}
\end{figure}

\section{Gaussian-process surrogate details}
\label{app:gpr_details}

For each Czochralski parameter sweep, Gaussian-process regression (GPR) is used to approximate the multi-output mapping
\begin{equation}
    (r,z,s)\mapsto(u_r,u_z,u_\theta,p,T),
\end{equation}
where $s$ denotes the active process parameter. The input coordinates and process parameter, as well as the target fields, are scaled before fitting. One complete operating condition is excluded from training and used as the case-wise holdout described in Section~\ref{sec:cfd_reference}.

The surrogate uses a fixed covariance kernel
\begin{equation}
    k(\mathbf{x},\mathbf{x}') = k_{\mathrm{RBF}}(\mathbf{x},\mathbf{x}')
    + k_{\mathrm{white}}(\mathbf{x},\mathbf{x}'),
\end{equation}
implemented as a unit-amplitude constant kernel multiplied by an anisotropic RBF kernel, with one initial length scale per input dimension, plus a white noise term. The RBF length scales and constant amplitude are fixed, and the white-noise level is set to $10^{-4}$. An additional diagonal regularization term of $\alpha=10^{-6}$ is used for numerical stability. Kernel hyperparameters are not optimized during fitting.

The GPR is trained jointly on the scaled output fields and returns both the predictive mean and predictive standard deviation. Predictions are transformed back to physical units using the target scaling parameters; predictive standard deviations are rescaled by the corresponding target standard deviations.

For each output field $y$, performance on the withheld operating condition is reported using RMSE, MAE, and standard-deviation-normalized RMSE,
\begin{equation}
    \mathrm{NRMSE} = \frac{\sqrt{\frac{1}{N}\sum_{i=1}^{N} (\hat{y}_i-y_i)^2}
    }{
        \sigma_y
    },
\end{equation}
where $\sigma_y$ is the standard deviation of that field in the holdout case. This normalization permits comparison across outputs with different physical units and characteristic scales.

Predictive uncertainty is additionally evaluated by computing the Pearson correlation between the absolute prediction error $|\hat{y}_i-y_i|$ and the GPR predictive standard deviation for each target. A positive correlation indicates that larger predictive uncertainty tends to coincide with larger holdout error. This quantity is used only as an error-awareness diagnostic and is not interpreted as a formal uncertainty calibration measure.

\end{document}